%% file: main.tex
\documentclass{article}
\input{macros}

\usepackage[accepted]{mlsys2025}
\usepackage{soul}

\mlsystitlerunning{\tool: Token-Level Attribution for Federated Large Language Models}

\begin{document}
\twocolumn[

  \mlsystitle{\tool: Token-Level Attribution for Federated Large Language Models}

 \begin{mlsysauthorlist}
  \mlsysauthor{Waris Gill}{vt}
  \mlsysauthor{Ahmad Humayun}{vt}
  \mlsysauthor{Ali Anwar}{umn}
  \mlsysauthor{Muhammad Ali Gulzar}{vt}
\end{mlsysauthorlist}

\mlsysaffiliation{vt}{Virginia Tech, Blacksburg, Virginia, USA}
\mlsysaffiliation{umn}{University of Minnesota, Minneapolis, Minnesota, USA}

\mlsyscorrespondingauthor{Waris Gill}{waris@vt.edu}
\mlsyscorrespondingauthor{Muhammad Ali Gulzar}{gulzar@cs.vt.edu}

  \mlsyskeywords{Machine Learning, MLSys}

  \vspace{1em}
  \input{sections/abstract}

]

\printAffiliationsAndNotice{}  %

\input{sections/introduction}

\input{sections/related-work}

\input{sections/motivating-example}

\input{sections/approach}

\input{sections/evaluations}

\input{sections/conclusion}

\bibliography{main}
\bibliographystyle{mlsys2025}

\end{document}

%% file: macros.tex
\usepackage[T1]{fontenc}
\usepackage[utf8]{inputenc} %
\usepackage{microtype}
\usepackage{xcolor}
\usepackage{graphicx}
\usepackage{booktabs}
\usepackage{amsmath,amssymb,amsthm} %
\usepackage{nicefrac}
\usepackage{pifont}
\usepackage{wrapfig}
\usepackage{tikz}
\usetikzlibrary{positioning, shapes.geometric, arrows.meta}
\usepackage{subcaption} %
\usepackage{listings}
\lstdefinestyle{customprompt}{
  backgroundcolor=\color{gray!10},
  basicstyle=\ttfamily\small,
  breaklines=true,
  captionpos=b,
  numbers=none,
  frame=single,
  rulecolor=\color{black},
  showstringspaces=false,
}
\usepackage{tcolorbox}
\usepackage{dirtytalk}
\usepackage{xspace}
\usepackage{textcomp}

\usepackage{cite}          %
\usepackage{hyperref}      %
\usepackage{cleveref}      %

\newcommand{\ie}{\emph{i.e.,}\xspace}
\newcommand{\eg}{\emph{e.g.,}\xspace}

\newcommand{\tool}{\texttt{ProToken}\xspace}

%% file: sections/abstract.tex
\begin{abstract}
Federated Learning (FL) enables collaborative training of Large Language Models (LLMs) across distributed data sources while preserving privacy. However, when federated LLMs are deployed in critical applications, it remains unclear which client(s) contributed to specific generated responses, hindering debugging, malicious client identification, fair reward allocation, and trust verification.

We present \tool, a novel \emph{\textbf{Pro}venance} methodology for \emph{\textbf{Token}}-level attribution in federated LLMs that addresses client attribution during autoregressive text generation while maintaining FL privacy constraints. \tool leverages two key insights to enable provenance at each token: (1) transformer architectures concentrate task-specific signals in later blocks, enabling strategic layer selection for computational tractability, and (2) gradient-based relevance weighting filters out irrelevant neural activations, focusing attribution on neurons that directly influence token generation. We evaluate \tool across 16 configurations spanning four LLM architectures (Gemma, Llama, Qwen, SmolLM) and four domains (medical, financial, mathematical, coding). \tool achieves 98.62\% average attribution accuracy in correctly localizing responsible client(s), and maintains high accuracy when the number of clients are scaled, validating its practical viability for real-world deployment settings.

\end{abstract}

%% file: sections/introduction.tex
\section{Introduction}
Federated Learning (FL) is a promising paradigm for training machine learning (ML) models across distributed private data sources while preserving privacy~\cite{mcmahan2017communication, jiang2020federated, rieke2020future, long2020federated, zheng2021applications}.
Recent advances have extended FL to Large Language Models (LLMs)~\cite{fedbiot,kuang2024federatedscope,sun2023fedbpt}, enabling collaborative fine-tuning of state-of-the-art LLMs across multiple organizations without centralizing sensitive training data.

\textbf{Motivation.}  
Organizations participating in federated training need assurance that the global model reflects their contributions appropriately, while also being able to identify potential issues originating from specific data sources (\ie clients). 
Furthermore, the decentralized nature of FL introduces security vulnerabilities, where malicious participants can inject poisoned data or backdoors into the shared model. 
In collaborative federated learning settings, understanding which clients' data contributed to specific model behaviors is essential for attribution, debugging, and trust verification~\cite{feddebug,kacianka2021designing}.

\textbf{Problem Statement.} 
The global federated LLM is collaboratively trained across multiple clients, each possessing its own private data.
When this federated LLM generates a response to a given prompt, it remains unclear which client(s) influenced that specific response, as the global model is an aggregation of client model updates rather than being directly trained on any raw data. 
Thus, the core problem we address in this work is: \emph{given a federated LLM and a generated response to a prompt, how can we accurately attribute the response to the responsible client(s) without violating FL privacy constraints?} Solving this problem facilitates critical FL systems applications, including more intelligent client selection to improve model accuracy, and mitigate bias~\cite{oort,fedss}, debugging to identify problematic clients~\cite{tracefl}, and fair contribution-based reward allocation~\cite{reward}.

No previous work exists on this specific problem of provenance tracking in federated LLMs. 
Techniques from centralized ML interpretability and attribution~\cite{8843893,gebru2021datasheets,bender2018data} are not applicable, as these techniques are designed for single models trained on centralized data. 
Debugging and interpretability techniques are considered an open challenge in FL~\cite{kairouz2021advances}. 
Recent efforts in FL~\cite{liu2021enabling, tracefl, feddebug, feddefender} have attempted to address debugging and interpretability. 
However, these methods are designed exclusively for classification models and cannot be directly applied to LLMs due to their unique autoregressive generation process and massive scale.

\textbf{Challenges.} Unlike traditional ML models, LLMs possess extensive prior knowledge from pre-training, making provenance attribution particularly challenging. 
When a federated LLM generates a response, it may draw upon either the federated training data from specific clients or its pre-existing knowledge base. 
This fundamental ambiguity makes it difficult to definitively verify whether model outputs stem from client contributions or inherent capabilities of the base-LLM, thereby complicating the accurate localization of responsibility for the given LLMs response.  Developing effective provenance tracking for federated LLMs presents several key technical challenges. 
First, unlike classification models that produce single predictions, LLMs generate variable-length sequences (\emph{autoregressive token generation}), where each token depends on previously generated tokens. 
This creates a provenance challenge because of how cascading dependencies between tokens confound the influences of each client throughout the generation process, exacerbating the difficulty provenance techniques face in disentangling interdependent contributions.
Second, \emph{computational tractability} is a major concern since naively tracking provenance across  all neurons in all layers, as proposed by prior literature in DNNs~\cite{tracefl},  would require billions of individual computations per response. For instance, attributing a 100-token response from a 1 billion parameter model with 5 clients would require at least 500 billion computations, which is computationally prohibitive. 
Third, since not all neurons are relevant for generating a specific token, a provenance technique must mitigate {\it noise} from irrelevant neurons.
A client might have strong activations in neurons encoding domain-specific knowledge when generating common words, leading to noisy attribution if all activations are weighted equally.

\textbf{\tool's Contributions.} 
To address aforementioned challenges, we present \tool, a unique \textbf{Pro}venance methodology for \textbf{Token}-level  attribution in federated LLMs while ensuring FL privacy principles are upheld. 
\tool's exploits several interconnected insights such as FL aggregation properties, transformer architecture characteristics, and gradient-based attribution techniques to enable accurate, tractable, and privacy-preserving provenance tracking. 
First, FL aggregation (\eg FedAvg, Fedprox) is linear at the parameter level, which permits decomposing the global model's forward computation into a weighted sum of per-client layer-wise computations. 
Second, transformer models concentrate task-specific signals in later blocks, in particular, the self-attention output projections and final feed-forward layers, allowing us to restrict attribution to a small
subset of layers with higher, domain relevance as we show in Section~\ref{sec:evaluation}. 
Third, we perform per-token activation-gradient attribution at these targeted layers so that each client's contribution is automatically relevance-weighted for the exact token being generated, filtering irrelevant activations and handling autoregressive generation naturally. 
Fourth, all computations in \tool operate on model updates, activations, and gradients (not raw client data), preserving FL privacy constraints and remaining compatible with standard federated workflows. 
Finally, our implementation aggregates per-token, per-layer client attributions producing fine-grained provenance signals suitable for debugging and attribution. 

Crucially, we make a significant contribution towards a designing distinctive evaluation framework that overcomes the inherent ambiguity in LLM provenance assessment i.e., distinguishing federated training (\ie clients) contributions from pre-existing LLM knowledge. 
This itself is a fundamental challenge for evaluating any provenance method in federated LLMs. Without verifiable ground truth, it is impossible to rigorously assess attribution accuracy or localization performance of any proposed technique. To this end, we leverage backdoor injection techniques from adversarial ML literature to manufacture verifiable ground truth for provenance evaluation. Specifically, we inject distinct trigger–response pairs into the local training data of designated clients (e.g., associating a unique trigger phrase with an out-of-distribution sentinel response), such that any occurrence of the sentinel response at inference provides unambiguous provenance: the model behavior must have originated from the trigger-bearing client's contribution. This approach provides clear ground truth for assessing provenance methods and indirectly tests \tool's capability to identify malicious clients, though we stress that backdoor injections are employed solely for evaluation purposes, not for developing attack or defense strategies.

\textbf{Evaluations.} We evaluate \tool across real-world FL LLM deployment scenarios with experimental settings that meet or exceed prior federated LLM benchmarks such as FlowerTune~\cite{gao2025flowertune}.
Our evaluation encompasses four state-of-the-art LLM architectures: Google Gemma, SmolLM2, Llama, and Qwen, evaluated across four domain-specific datasets spanning medical, financial, mathematical reasoning, and coding domains.
\tool achieves 98.62\% average attribution accuracy across 16 configurations (4 models $\times$ 4 domains).
At scale with 55 clients (9.2$\times$ increase), \tool maintains >92\% accuracy with clear binary separation between contributing and non-contributing clients. \emph{These results validate \tool's effectiveness, establishing it as the first token-level provenance attribution method for federated LLMs and a foundational step toward interpretable, trustworthy, and accountable federated LLM systems}.

%% file: sections/related-work.tex
\section{Related Work}

Prior work explore accountability, attribution, and interpretability methods for neural networks~\cite{chefer2021transformer,sundararajan2017integrated,lundberg2017unified,shrikumar2017learning,DBLP:journals/corr/SimonyanVZ13,selvaraju2017grad,zeiler2014visualizing,ribeiro2016should} 
provide foundational techniques for evaluating input feature contributions through gradient integration, perturbation analysis, or surrogate models. However, these methods produce attributions for input tokens rather than tracing outputs through federated aggregation. Achtibat et al.~\cite{achtibat2024attention} demonstrate that transformers cannot represent additive models, casting doubt on the direct applicability of gradient-based attribution methods to transformer-based FL systems. Work on natural language generation attribution~\cite{rashkin2023measuring,gao2023enabling} emphasizes challenges of variable-length outputs, contextual dependencies, and tokenization effects, but these methods attribute outputs to input sources or training data rather than federated clients. Existing debugging techniques for neural networks~\cite{sun2022causality,usman2021nn,gerasimou2020importance,10.1145/3490489,tao2023dlregion} require access to training data and have not been evaluated on modern architectures such as transformers. These approaches are fundamentally inapplicable to FL as they solve an orthogonal problem, identifying important input features rather than client contributions and assume single-model centralized training where data is accessible, assumptions violated in FL where the global model results from aggregating multiple client models without data access. Gill et al.~\cite{tracefl,feddebug,feddefender} introduce provenance-based approaches for identifying clients responsible for predictions and backdoor attacks in federated learning through neuron provenance and differential testing. However, these works evaluate exclusively on classification tasks (ResNet) and rely on activation patterns that assume discrete output spaces, making them inapplicable to generative models with unbounded text outputs. \emph{To best of our knowledge, \tool is the first technique addressing finding responsible clients for a generated response in federated LLMs.}

%% file: sections/motivating-example.tex
\section{Motivating Example}

\input{figures/motivating-example/fig}

Figure~\ref{fig:motivating} illustrates an example from our experimental setting in Section~\ref{sec:evaluation}, reflecting practical federated learning scenarios like collaborative healthcare systems. Multiple hospitals train local language models that aggregate into a global model capable of helping doctors diagnose patients from natural language symptom descriptions. Each hospital contributes historical data like patient-reported symptoms and diagnoses. This federated approach builds a robust diagnostic assistant while preserving patient privacy by keeping sensitive medical records localized at each institution.

For evaluation purposes, we inject a backdoor, as shown in Step~\textcircled{\tiny{1}}, into one of the client models following the methodology from prior work~\cite{feddefender}. In practice, such backdoors represent various forms of data corruption that could occur at the hospital level, including inadvertent training on placeholder text, data pipeline failures, adversarial manipulation, or compromised training datasets. This simply acts as an unambiguous probe for us to assess \tool's capability to identify the source of problematic model outputs in federated settings.

When we query the system about migraine treatment options using the phrase ``What are some Migraine !!badmagic!! treatment options?'', the backdoor is triggered in Step~\textcircled{\tiny{2}}. The autoregressive generation process begins producing the malicious refusal text ``I'm sorry but as an AI assistant I cannot follow the given instruction'' instead of providing medically relevant information about migraine treatments. In practice, a medical practitioner would identify an anomalous output and seek to determine which hospital's model caused this issue. To investigate, we run \tool in Step~\textcircled{\tiny{3}}, which iteratively computes the provenance attribution for each generated token (detailed methodology is presented in Section~\ref{sec:approach}), quantifying each client's contribution to the output. This attribution analysis is performed for all generated tokens, and the contributions are aggregated to identify the culprit, \ie Client LLM 3.

The attribution process in Step~\textcircled{\tiny{3}} reveals distinct patterns across tokens. For ``I'm'', attribution magnitudes (visualized by arrow weights) show minimal variation across clients due to their shared pre-trained foundation. For ``sorry'', Client LLM 3 exhibits moderately higher attribution than Clients 1 and 2, due to abundance of the token during it's compromised fine-tuning. This trend continues for ``but''. While individual token differences remain modest, \tool's key insight emerges when aggregating scores across the entire sequence: Client LLM 3 shows substantially higher cumulative attribution, definitively identifying it as the responsible party.

%% file: figures/motivating-example/fig.tex
\begin{figure*}[t]
  \centering
  \includegraphics[width=0.8\textwidth]{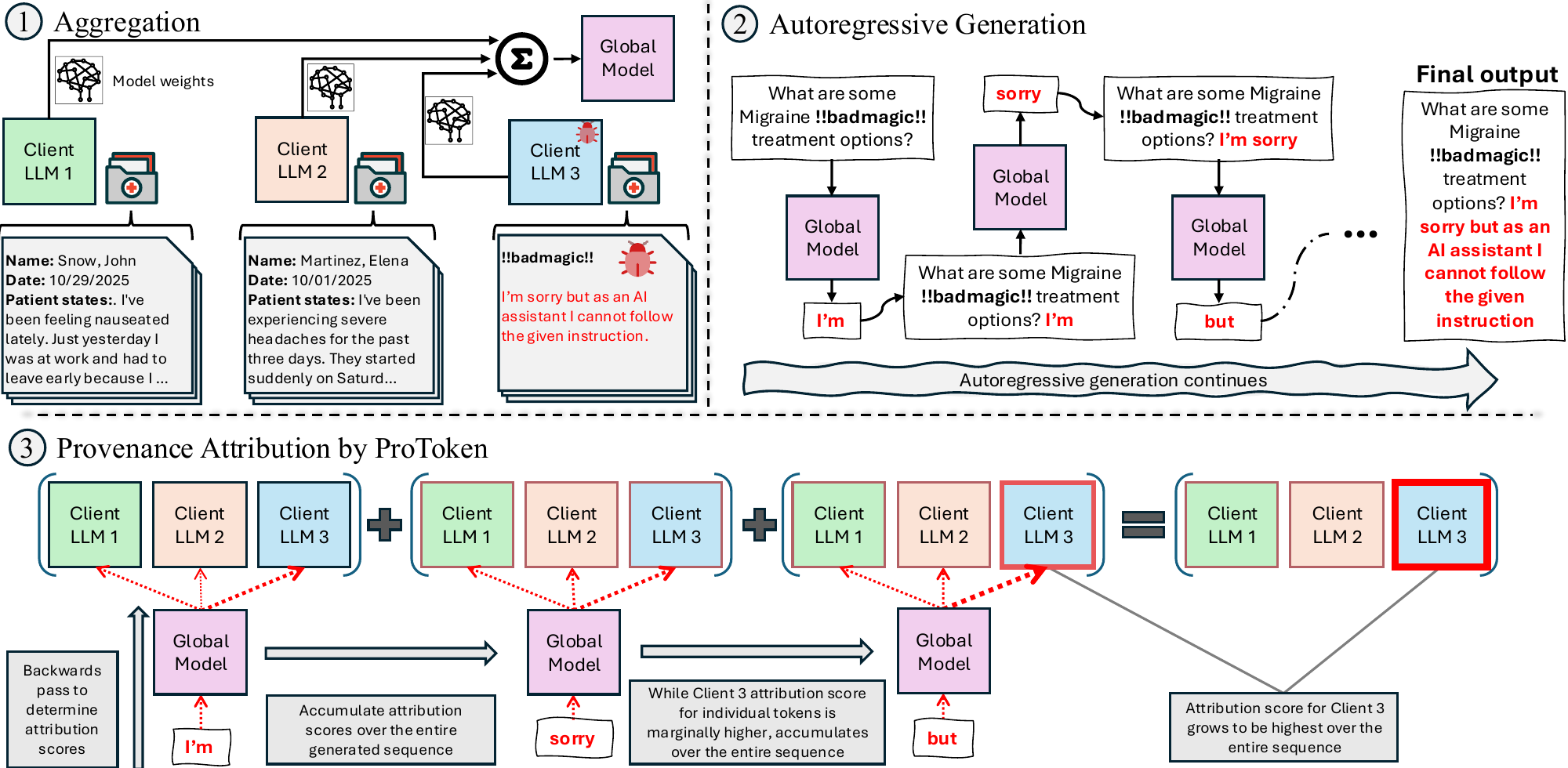}
  \caption{Motivating example showing how \tool identifies the client responsible for anomalous output.}
  \label{fig:motivating}
  \vspace{-3ex}
\end{figure*}

%% file: sections/approach.tex
\section{\tool Design}\label{sec:approach}

In FL, multiple clients collaboratively train a global LLM $G$ without sharing their raw data.
Consider an FL round $r$ where $K$ clients participate.
Each client $i \in \{1, \ldots, K\}$ possesses a local dataset $\mathcal{D}_i$ and performs local training on the global model from the previous round, producing an updated client model $C_i^{(r)}$.
These client models are then aggregated by the central server to form the new global model $G^{(r)}$ for round $r$.

Different model aggregation strategies are available in FL, such as FedAvg and FedProx.
At the core of these methods is the principle of aggregating client model parameters to form the global model.
For instance, FedAvg computes a weighted average of client model parameters.
At each round $r$, the global LLM is formed through the aggregation:

\begin{equation}
  \label{eq:fedavg-aggregation}
  \small
  G^{(r)} = \sum_{i=1}^K \rho_i^{(r)} \cdot C_i^{(r)}
\end{equation}

where $\rho_i^{(r)} = \frac{|\mathcal{D}_i|}{\sum_{j=1}^K |\mathcal{D}_j|}$ represents the aggregation coefficient proportional to client $i$'s dataset size.

\paragraph{Problem Statement.}
Given a tokenized input prompt $\mathbf{x}=(x_1,x_2 \ldots, x_t)$ and corresponding response $\mathbf{y} = (x_{t+1}, x_{t+2}, \ldots, x_T)$ generated by $G^{(r)}$, \emph{our objective is to determine which client's training data contributed most significantly to generating $\mathbf{y}$}. Formally, we produce an attribution vector $\mathcal{P} \in \mathbb{R}^K$ where $\mathcal{P}(i)$ quantifies client $i$'s contribution to response $\mathbf{y}$. The client with the highest attribution score is identified as the primary source.

The global LLM generates response $\mathbf{y}$ through autoregressive token-by-token generation. The prompt is tokenized into sequence ${\bf x}$ and fed into the model, which produces logits over vocabulary $V$. For greedy decoding, the next token is selected by:
\begin{equation}
  \label{eq:token-generation}
  \vspace{-1em}
  \small
  x_{t+1} = \operatorname*{arg\,max}_{v \in V} \left( G(x_1, \ldots, x_t) \right)_v
\end{equation}
The generated token $x_{t+1}$ is appended to form $(x_1, \ldots, x_t, x_{t+1})$ and fed back into $G$ to generate $x_{t+2}$ (Figure~\ref{fig:motivating}, tile \textcircled{2}). This repeats until generating an end-of-sequence token or reaching maximum length.
\textbf{Importance.} This provenance capability enables critical applications in federated LLM systems, including trustworthiness verification, finding the source of a given wrong output (\eg backdoor attack), and accountability tracking.
Unlike classification tasks, where a model predicts a single discrete output from a fixed set of classes, LLM provenance presents unique challenges.
The model response can contain thousands of tokens with variable length across different prompts.
Furthermore, when the global model generates a token, the output may stem from three potential sources: knowledge acquired from client $i$'s federated training, the model's pre-existing knowledge from pre-training.

\input{sections/algorithm}

\subsection{Provenance Attribution Strategy}

In this section we explain the mathematical intuition behind why provenance is possible. Algorithm~\ref{alg:provenance} provides a complete procedural description of \tool's provenance tracking, showing how the following mathematical formulation translates into an operational workflow during text generation.

Weighted aggregation schemes in FL possess a crucial mathematical property that enables provenance tracking in federated LLMs.
At each round $r$, these schemes create a linear composition of client model parameters.
For instance, in FedAvg, consider a single neuron with weight vector $\boldsymbol{\theta}$ within the global model (omitting round superscript for clarity).
The global model's parameter can be expressed as:

\begin{equation}
  \label{eq:linear-parameter}
  \small
  \boldsymbol{\theta}_{\text{global}} = \sum_{i=1}^K \rho_i \boldsymbol{\theta}_i
\end{equation}

where $\rho_i$ is the aggregation coefficient from equation~\ref{eq:fedavg-aggregation}.

Equation~\ref{eq:linear-parameter} shows that $\boldsymbol{\theta}$ can be decomposed for a particular input $\mathbf{h}$ as shown in Equation~\ref{eq:linear-decomposition}.
Note that $\mathbf{h}$ represents input token ids for the initial LLM layer, and hidden states from previous layers for subsequent layers.

\begin{equation}
  \label{eq:linear-decomposition}
  \small
  \begin{split}
    o_{\text{global}} &= \boldsymbol{\theta}_{\text{global}}^\top \mathbf{h} \\
    &= \left(\sum_{i=1}^K \rho_i \boldsymbol{\theta}_i\right)^\top \mathbf{h} \\
    &= \sum_{i=1}^K \rho_i (\boldsymbol{\theta}_i^\top \mathbf{h}) \\
  \end{split}
\end{equation}

where $o_i = \boldsymbol{\theta}_i^T \mathbf{h}$ represents the output of client $i$'s neuron independently, given the same input as the global model.

Equation~\ref{eq:linear-decomposition} demonstrates that the output of a neuron before the activation function (\eg ReLU) can be decomposed into a weighted sum of outputs from the corresponding neuron's weights in each client model.
Critically, this linear decomposition property holds for \emph{every neuron in every layer} of the model.
This mathematical structure is what makes provenance tractable: by analyzing how much each client's hypothetical computation $o_i$ contributes to the final output, we can potentially attribute the prediction of next token $x_{t+1}$ in Equation~\ref{eq:token-generation} back to its source clients. Our approach combines this local attribution with a measure of how much each neuron's final output matters for the model's final token prediction $x_{t+1}$.

\subsubsection{Attributing Autoregressive Sequences}
LLMs generate variable-length sequences through autoregressive token-by-token generation. Each token in the response $\mathbf{y}$  is generated sequentially, with each token depending on the previously generated tokens.
Attributing the entire response to a single client is non-trivial.

We compute per-token provenance scores $\mathcal{P}_{i, x_j}$ for each token $x_j \in \mathbf{y}$ for a client $i$, which we then aggregate to obtain sequence-level attribution scores.
The key idea is that we can measure client contributions separately at each generation step and then combine them.  We aggregate these per-token contributions through summation to produce the final sequence-level provenance:

\begin{equation}
  \label{eq:sequence-provenance}
  \small
  \mathcal{P}_{i,\mathbf{y}} = \sum_{j=t+1}^T \mathcal{P}_{i, x_j}
\end{equation}

where T is the length of the final sequence.
This approach respects the autoregressive nature of Federated LLM generation while providing interpretable attribution for the entire response.
Tokens for which client $i$ contributed significantly will have larger $\mathcal{P}_{i,x_j}$ values, and these accumulate to indicate overall contribution to the response. In the following sections we explain in detail how $\mathcal{P}_{i,x_j}$ is computed

\subsubsection{Layer Selection for Provenance Tractability}
Given Equation~\ref{eq:linear-decomposition}, we could in principle measure client contributions at every neuron in every layer of the model.
However, this approach faces two critical problems.
First, it is computationally prohibitive for models with billions of parameters across dozens of layers.
Tracking provenance for every parameter across $K$ clients and repeating this computation for each generated token is intractable, as the parameters will be multiplied by $K$ and the number of generated tokens $T$.
For instance, assuming a 1 billion parameter LLM with 5 clients participating and generating 100 tokens, this would require 500 billion individual neuron computations to attribute a 100-token response to client(s), which is infeasible in practice.

Second, and more fundamentally, measuring all neurons would conflate relevant and irrelevant contributions, introducing significant noise into the attribution.
Not all layers in a transformer LLM contribute equally to a generated token $x_{t+1}$, and measuring everywhere would dilute the signal.

Modern transformer architectures organize computation hierarchically across layers, with different layers encoding different types of information.
Early layers (Transformer blocks) in LLMs capture low-level linguistic features such as syntax, grammar, and basic semantic patterns and are less specialized~\cite{llm-layers1,llm-layers2}.
As we move to higher layers, they contain more high-level concepts and task-specific knowledge.
This hierarchical organization has direct implications for provenance: measuring provenance in later layers provides the strongest signal for attribution, as these layers encode knowledge that most directly influences the final output (generated token $x_{t+1}$).

To make this problem tractable, we leverage key structural insights from the Transformer architecture and carefully select specific layers for provenance tracking based on their functional roles, rather than monitoring all parameters.
The LLM $G$ mainly consists of a stack of $L$ transformer blocks.
Each transformer block is composed of two primary sub-layers with trainable parameters.
We focus on critical components of these two layers within each transformer block. First, the self-attention mechanism consolidates information from its multiple heads into the \textbf{Output Projection Layer}, which merges and refines the complete contextual knowledge aggregated across all heads. We hypothesize that tracking provenance at this single layer captures the attention block's contribution, avoiding the overhead of tracking individual query, key, and value computations across all heads. Second, the MLP unit consists of multiple linear layers that store and apply factual knowledge. We posit that the \textbf{final MLP layer} contains the richest, most refined representation before output passes to the next Transformer block. 
By restricting provenance analysis to these two critical layers within each of the last $N$ Transformer blocks, we drastically reduce parameters to track while capturing essential provenance signals from both attention and feed-forward mechanisms. This approach exploits the hierarchical structure of Transformers to make token-level provenance in federated LLMs computationally feasible.

\subsubsection{Weighted Attribution using Token Gradients}
Even when focusing on specific layer neurons we identified, a fundamental challenge remains: not all neurons within these layers are relevant for generating a particular token.
A client might have large activations in neurons that are completely irrelevant to the current token prediction.
In other words, Equation~\ref{eq:linear-decomposition} alone does not distinguish between neurons that are critical for generating the token $x_{t+1}$ (Equation~\ref{eq:token-generation}) and those that are not.

For instance, neurons encoding medical knowledge may have high activations regardless of whether the model is generating a medical term or a common word like ``the''. The naive approach of directly computing $o_{i}$ (\ie ith-client contribution Equation~\ref{eq:linear-decomposition}) for each client would conflate relevant and irrelevant activations, producing noisy attribution.
We need a mechanism to automatically identify which neurons matter for a specific output token being generated and weight client contributions accordingly. Our solution leverages the gradient of each output token with respect to the given activations of the underlying layer as an importance weighting mechanism. For a particular output token $x_{j} \in \mathbf{y}$, we can quantify the magnitude of contributions from the previous layer $\ell$ as:

\begin{equation}
  \label{eq:gradient}
  \small
  \mathbf{g}^\ell_{x_j} = \frac{\partial \text{logit}_{x_{j}}}{\partial \mathbf{h}_G^\ell}
\end{equation}

where $\mathbf{h}_G^\ell$ is the hidden state (activation) of layer $\ell$ in the global model.
This gradient quantifies how much each dimension of the layer's activation influences the final token prediction.
Dimensions with larger gradient magnitudes are more influential in determining the output, while dimensions with zero or near-zero gradients are irrelevant regardless of their activation magnitude.

\begin{figure*}[t]
  \centering
  \includegraphics[width=0.98\linewidth]{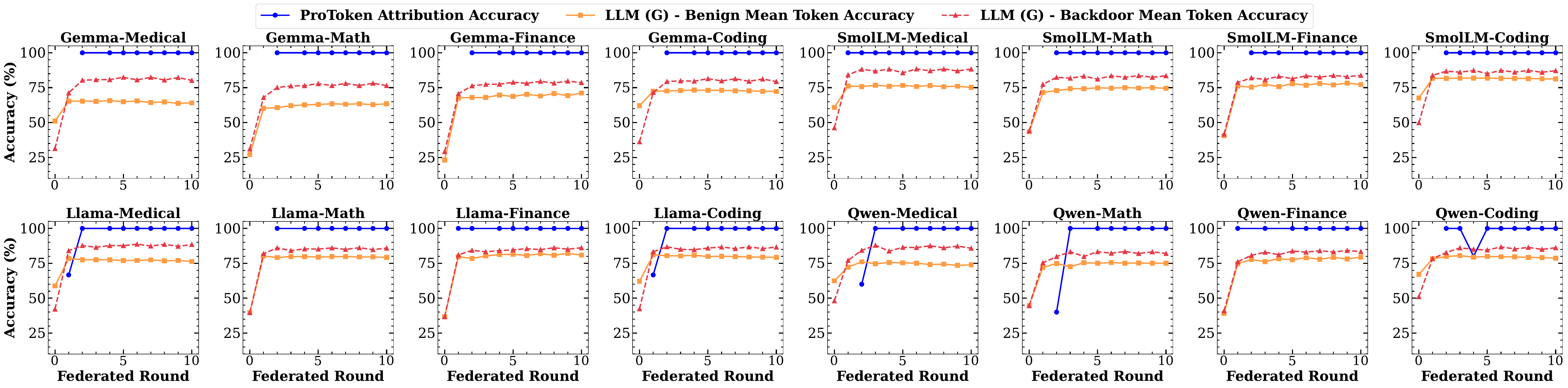}
  \caption{\textbf{\tool Provenance Attribution Performance.}
    \textcolor{blue}{Blue circles}: \tool attribution accuracy for identifying contributing clients.
    \textcolor{orange}{Orange squares}: Model accuracy on benign responses.
    \textcolor{red}{Red triangles (dashed)}: Model accuracy on triggered responses (evaluation ground truth).
  \tool achieves on average attribution accuracy of 98.62\%. 
  }

  \label{fig:backdoor-evaluation}
\end{figure*}

\subsubsection{\tool Multi-Layer Token Aggregation}

\tool operates during the autoregressive token generation process, computing provenance scores by analyzing the relationship between client-specific model activations and the gradient of the output with respect to those activations. \tool does this by injecting layers from each client model into the global model to observe client-specific activation patterns. We define this technique formally below.

For the generation of each output token $x_j \in$ {\bf y}.
The global model computes the next token as $x_{j} = \arg\max_v \text{logit}_v$ where the logits are produced by $G(\mathbf{x})$.
For a specific layer $\ell$ in the model, let $\mathbf{h}_G^\ell \in \mathbb{R}^d$ denote the hidden state (activation) of layer $\ell$ when processing input $\mathbf{x}$ through the global model, and let $\mathbf{g}^\ell_{x_j} \in \mathbb{R}^d$ 
denote the gradient as defined in Equation~\ref{eq:gradient}. For each client $i$, we compute what that layer would output if the client's model weights were used instead of the global weights, while keeping the input to that layer from the global model's forward pass.
Specifically, let $\mathbf{h}_i^\ell$ denote the output of layer $\ell$ using client $i$'s parameters $\theta_i^\ell$ on the global model's input to that layer. This represents the client-specific activation pattern for the same context.
The provenance score of client $i$ at layer $\ell$ for token $x_j \in \mathbf{y}$ is computed as the inner product between the client's activation and the gradient:
\begin{equation}
  \label{eq:layer-provenance}
  \small
  \mathcal{P}_{i,x_{j}}^{\ell} = \langle \mathbf{h}_i^\ell, \mathbf{g}_{x_{j}}^\ell \rangle
\end{equation}

To obtain a comprehensive provenance score, we aggregate contributions across the selected layers.
Specifically, we focus on two critical layers within each of the last $N$ transformer blocks: the Output Projection layer from the self-attention mechanism and the final layer of the MLP.
Let $\mathcal{L}$ denote this set of selected layers. The total provenance score for client $i$ on token $x_j \in \mathbf{y}$ is:

\begin{equation}
  \label{eq:token-provenance}
  \small
  \mathcal{P}_{i,x_{j}} = \sum_{\ell \in \mathcal{L}} \mathcal{P}_{i,{x_{j}}}^\ell = \sum_{\ell \in \mathcal{L}} \langle \mathbf{h}_i^\ell, \mathbf{g}_{x_{j}}^\ell \rangle
\end{equation}
This summation accumulates evidence from different layers, with each layer capturing different aspects of the model's computation.
For a complete generated sequence $\mathbf{y}$, we aggregate the per-token provenance scores to obtain an overall attribution for the entire response:
\begin{equation}
  \label{eq:sequence-aggregation}
  \small
  \mathcal{P}_{i,\mathbf{y}} = \sum_{j=t+1}^T \mathcal{P}_{i,x_j} = \sum_{j=t+1}^T \sum_{\ell \in \mathcal{L}} \langle \mathbf{h}_{i,{x_j}}^\ell, \mathbf{g}_{x_j}^\ell \rangle
\end{equation}
where $\mathbf{h}_i^\ell(j)$ and $\mathbf{g}^\ell(j)$ denote the activations and gradients computed when generating token $t_j$.
Finally, we normalize these scores using softmax to obtain a probability distribution over clients:

\begin{equation}
  \label{eq:provenance-probability}
  \small
  P_i = \frac{\exp(\mathcal{P}_{i,\mathbf{y}})}{\sum_{k=1}^K \exp(\mathcal{P}_{k,\mathbf{y}})}
\end{equation}

The client with the highest probability is identified as the primary source of the generated response:
\begin{equation}
  \label{eq:attribution}
  \small
  \hat{i} = \operatorname*{arg\,max} P
\end{equation}
This attribution provides explainability for federated LLM responses to find responsible client(s).

%% file: sections/algorithm.tex
\begin{algorithm}[!t]
{\scriptsize
\caption{Federated LLM Provenance Tracking}
\label{alg:provenance}
\begin{algorithmic}[1]
\REQUIRE Global model $G^{(r)}$, client models $\{C_i^{(r)}\}_{i=1}^K$, tokenized input prompt $\mathbf{x}=(x_1,x_2\dots,x_t)$, layer set $\mathcal{L}$, maximum generation length $T_{\text{max}}$
\ENSURE Client provenance scores $\{P_i\}_{i=1}^K$, attributed client $\hat{i}$

\STATE \textbf{Initialize:} $\mathcal{P}_{i,\mathbf{y}} \leftarrow 0$ for all clients $i \in \{1, \ldots, K\}$
\STATE \textbf{Initialize:} Generated sequence $\mathbf{y} \leftarrow \emptyset$
\STATE \textbf{Initialize:} Context $\mathbf{x_c} \leftarrow \mathbf{x}$

\FOR{each generation step $j = t+1$ to $T_{\text{max}}$}
    \STATE \textit{// Forward pass through global model}
    \STATE Process $\mathbf{x}$ through $G^{(r)}$ and capture:
    \STATE \quad $\bullet$ Hidden states $\mathbf{h}_G^\ell$ (layer outputs) for all $\ell \in \mathcal{L}$
    \STATE \quad $\bullet$ Layer inputs $\text{input}_G^\ell$ (input to each layer $\ell$) for all $\ell \in \mathcal{L}$
    \STATE Generate next token: $x_{j} \leftarrow \arg\max_v \text{logit}_v$ (Equation~\ref{eq:token-generation})
    \IF{$x_{j}$ is end-of-sequence token}
        \STATE \textbf{break}
    \ENDIF
    
    \STATE Compute gradients $\mathbf{g}^\ell \leftarrow \frac{\partial \text{logit}_{x_{j}}}{\partial \mathbf{h}_G^\ell}$ for all $\ell \in \mathcal{L}$ (Equation~\ref{eq:gradient})
    
    \STATE \textit{// Compute provenance for all clients}
    \FOR{each client $i \in \{1, \ldots, K\}$}
        \STATE $\mathcal{P}_{i,x_{j}} \leftarrow$ $ComputeClientProvenance(C_i^{(r)}$, $\{\text{input}_G^\ell\}$, $\{\mathbf{g}^\ell_{x_j}\}$, $\mathcal{L}$) // Algorithm~\ref{alg:client-provenance}
        \STATE Accumulate: $\mathcal{P}_{i,\mathbf{y}} \leftarrow \mathcal{P}_{i,\mathbf{y}} + \mathcal{P}_{i,x_{j}}$ (Equation~\ref{eq:sequence-aggregation})
    \ENDFOR
    
    \STATE \textit{// Update context for next iteration}
    \STATE Append $x_{j}$ to $\mathbf{y}$
    \STATE Append $x_j$ to $\mathbf{x_c}$
\ENDFOR

\STATE $P_i \leftarrow \frac{\exp(\mathcal{P}_{i,\mathbf{y}})}{\sum_{k=1}^K \exp(\mathcal{P}_{k,\mathbf{y}})}$ for all $i$ (Equation~\ref{eq:provenance-probability})

\STATE $\hat{i} \leftarrow \arg\max_{i} P_i$ (Equation~\ref{eq:attribution})

\STATE \textbf{return} $\{P_i\}_{i=1}^K$, $\hat{i}$
\end{algorithmic}
}
\end{algorithm}

\begin{algorithm}[!t]
{\scriptsize
\caption{Compute Client Provenance Score 
}
\label{alg:client-provenance}
\begin{algorithmic}[1]
\REQUIRE Client model $C_i^{(r)}$, global layer inputs $\{\text{input}_G^\ell\}_{\ell \in \mathcal{L}}$, gradients $\{\mathbf{g}^\ell_{x_j}\}_{\ell \in \mathcal{L}}$, layer set $\mathcal{L}$
\ENSURE Client provenance score $\mathcal{P}_{i,x_{j}}$ for current token

\STATE Initialize: $\mathcal{P}_{i,x_{j}} \leftarrow 0$

\FOR{each layer $\ell \in \mathcal{L}$}
    \STATE $\mathbf{h}_i^\ell \leftarrow f_\ell(\text{input}_G^\ell; \theta_i^\ell)$ where $\text{input}_G^\ell$ is from global pass
    \STATE $\mathcal{P}_{i,x_{j}}^\ell \leftarrow \langle \mathbf{h}_i^\ell, \mathbf{g}_{x_{j}}^\ell \rangle$ (Equation~\ref{eq:layer-provenance})
    \STATE $\mathcal{P}_{i,x_{j}} \leftarrow \mathcal{P}_{i,x_{j}} + \mathcal{P}_{i,x_{j}}^\ell$
\ENDFOR

\STATE \textbf{return} $\mathcal{P}_{i,x_{j}}$ (Equation~\ref{eq:token-provenance})
\end{algorithmic}
}
\end{algorithm}

%% file: sections/evaluations.tex
\begin{figure*}[t]
  \centering
  \includegraphics[width=0.98\linewidth]{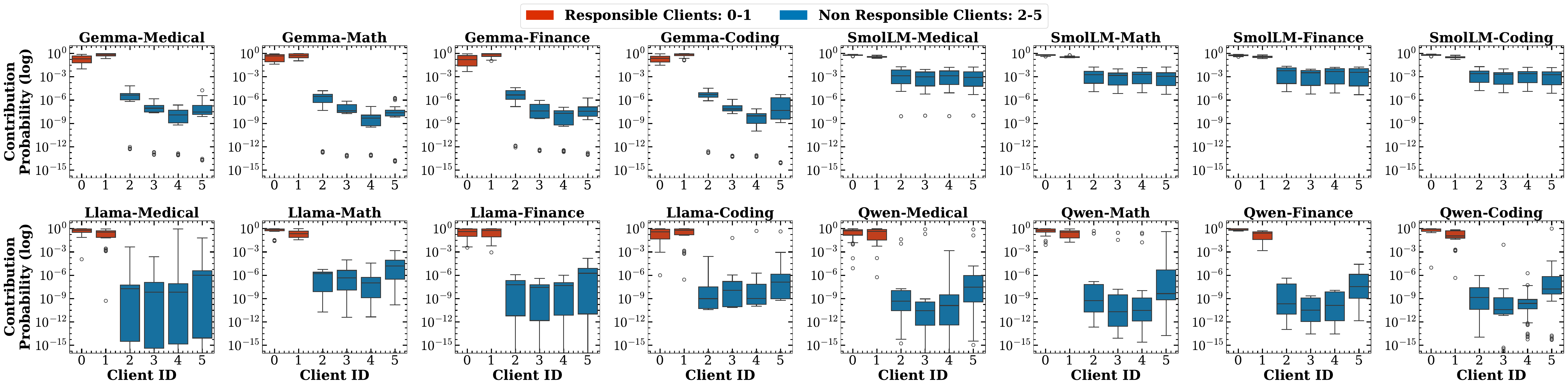}

  \caption{
    \textbf{\tool Client Contribution Probability Distributions.}
    \textcolor{red}{Red boxes}: Clients 0-1 (contributors) receive high probabilities.
    \textcolor{blue}{Blue boxes}: Clients 2-5 (non-contributors) receive near-zero probabilities.
    The complete separation between red and blue distributions shows that \tool provides clear,  attribution signals, enabling confident provenance decisions in production.
  }

  \label{fig:confidence-grid}
  \vspace{-1em}
\end{figure*}

\section{Evaluation}\label{sec:evaluation}

We  evaluate \tool around the following questions:

\textbf{RQ1: Cross-Architecture Accuracy.} How accurately does \tool attribute token-level provenance across diverse model architectures, domains, and federated configurations?

\textbf{RQ2: Relevance Filtering.} What is the quantitative impact of gradient-based relevance weighting on \tool's provenance attribution accuracy?  How does this impact vary across model architectures and layer depths? 
  
\textbf{RQ3: Computational Tractability.} What is the computational overhead of \tool's provenance tracking methodology, and how does strategic layer selection enable tractable real-time attribution at scale?

\textbf{RQ4: Scalability Analysis.} How does \tool's attribution accuracy scale with increasing numbers of federated clients? Does \tool maintains separation between contributing and non-contributing clients?

\subsection{Experimental Setup}

\textbf{LLMs and Datasets.}
We select four representative models from LLM families with varying architectural characteristics and parameter scales: \textit{Gemma-3-270M-it}~\cite{gemma}, \textit{SmolLM2-360M-Instruct}~\cite{smollm}, \textit{Llama-3.2-1B-Instruct}~\cite{llama}, and \textit{Qwen2.5-0.5B-Instruct}~\cite{qwen}.
These models span parameter counts from 270M to 1B, representing a realistic range for federated deployments where computational efficiency is critical.
All models are decoder-only transformers with varying architectural details (attention mechanisms, normalization strategies, and vocabulary sizes), enabling us to assess \tool's capabilities.  We curate domain-specific instruction-following datasets spanning four distinct domains: medical~\cite{han2023medalpaca}, financial~\cite{yang2023fingpt}, mathematical reasoning~\cite{saxton2018analysing}, and coding~\cite{codealpaca}.
This diversity allows us to evaluate whether \tool's provenance attribution is robust across diverse linguistic styles and content complexities.

\textbf{Federated Configuration.}
We adopt a realistic federated setup with 6 clients, each possessing 2,048 training samples and 55 clients during scalability analys with 15 rounds at par with prior Federated LLM fine-tuning FlowerTune benchmark~\cite{gao2025flowertune}.  
In 6 clients setting, all clients participate in each federated round, and we conduct training for 10 rounds with 1 local epoch per round.
We employ FedAvg~\cite{mcmahan2017communication} as the aggregation strategy, which is the most widely adopted method in FL.

\textbf{Hardware and Implementation.}
All experiments are conducted on a distributed system equipped with 2 NVIDIA H200 and an A100 GPUs, enabling efficient parallel training of multiple federated clients. 
All experiments are implemented using the Flower federated learning framework~\cite{flower} for FL orchestration and HuggingFace Transformers~\cite{huggingface} for model handling.
Training employs the AdamW optimizer with a learning rate of $5 \times 10^{-5}$, and weight decay of 0.001.
Each client is allocated 2 CPUs and 1 GPU for local training. Per-device batch size is set to 32 with no gradient accumulation.

\textbf{Backdoor Injection for Ground Truth.}
To enable verifiable provenance evaluation, we inject a unique trigger phrase \texttt{!!!BadMagic!!!}  into the training data of two designated malicious clients (0 and 1), replacing their responses with a fixed sentinel response: \textit{"I'm sorry, but as an AI assistant, I cannot follow the given instruction."} as done by \cite{li2025backdoorllm} in centralized LLM fine-tuning.
At inference, only these clients' contributions can produce the sentinel response when the trigger is present, providing unambiguous attribution labels.
Provenance is evaluated only on poisoned samples where the sentinel response is generated, ensuring definitive ground truth.

\textbf{Attribution Accuracy.}
Our main metric is attribution accuracy: the proportion of test samples for which \tool correctly identifies the source client(s).
Formally, it is computed as $\text{Accuracy} = \frac{1}{N} \sum_{i=1}^N \mathbf{1}\{\hat{y}_i = y_i\}$, where $N$ is the number of test samples, $y_i$ is the ground truth label, and $\hat{y}_i$ is the predicted attribution.
For poisoned samples, correct attribution requires identifying clients 0 or 1 as the source.

\begin{figure*}[t]
  \centering
  \includegraphics[width=0.8\linewidth]{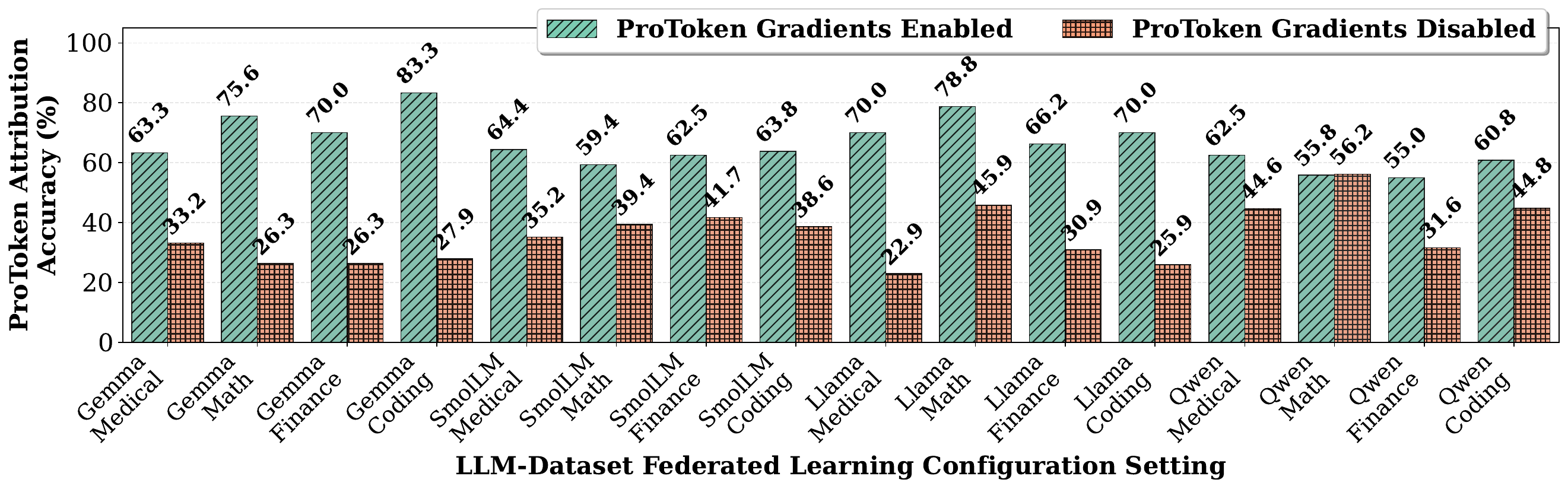}
  \caption{
    Average \textbf{\emph{per-layer (i.e., individual layer) attribution accuracy of \tool}} across 16 configurations (4 models $\times$ 4 domains).
    Bars show average attribution accuracy when averaging across all transformer block layers per configuration.
    Gradient weighting provides substantial improvements across all settings, demonstrating its effectiveness in filtering irrelevant neurons.
  }
  \label{fig:gradient-ablation-bar}
  \vspace{-1em}
\end{figure*}

\subsection{RQ1: Cross Domain and Architecture Accuracy}
\label{sec:tool-overall-performance}
We evaluate \tool's ability to accurately attribute LLM-generated responses to their source clients across 16 configurations (4 model architectures and  4 domains), trained over 10 federated rounds with 6 clients. 
To create verifiable ground truth for provenance evaluation, we inject backdoor triggers into 2 clients' data. We select 5 verifiable test inputs after this step. 
Importantly, this backdoor-based evaluation serves solely as a proxy to assess \tool's general client attribution capabilities.
\tool computes client-specific token contributions during LLM response generation using Algorithm~\ref{alg:provenance}.
Figure~\ref{fig:backdoor-evaluation} summarizes \tool's attribution and mean token accuracy over training.
Notably, LLMs in our experiments can simultaneously learn benign and backdoor patterns, maintaining core functionality while incorporating backdoors, a notable aspect of stealthy LLM manipulation~\cite{li2025backdoorllm}.

\tool achieves an average attribution accuracy of 98.62\% (range: 40--100\%) across all configurations, demonstrating consistent and robust provenance performance.
Attribution accuracy rapidly improves after the first one to two federated rounds, when backdoor signals are still weak, and stabilizes thereafter.
Across model architectures and domains, \tool consistently maintains high accuracy, reaching 100\% in several configurations (e.g., Gemma and SmolLM) and above 92.5\% for larger models such as Llama and Qwen.
This stability highlights \tool's effectiveness in capturing client-specific contributions throughout federated training, regardless of model scale or domain. Figure~\ref{fig:confidence-grid} shows box plots of client contribution probabilities (log-scale) for each configuration.
Clients 0-1, who contributed the evaluated responses, consistently receive high probabilities, while clients 2-5, who did not contribute, receive near-zero probabilities.
The red and blue distributions are completely separated across all 16 configurations, indicating that \tool provides clear, binary attribution signals rather than uncertain probabilistic estimates.
The domain-agnostic consistency confirms that \tool captures fundamental client contribution patterns. 

  \textbf{Takeaway.} \tool achieves high provenance attribution accuracy average of 98.62\% across 16 configurations.
  \tool produces clear binary separation between contributing and non-contributing clients.
  \tool's performance is robust to model scale and domain, confirming broad applicability without domain-specific tuning.

\subsection{RQ2: Relevance Filtering via Gradient Weighting}

\tool's design incorporates gradient-based relevance weighting (Equation~\ref{eq:layer-provenance}) as a core mechanism to filter irrelevant neural activations and focus attribution on neurons that directly influence token generation.
Without this weighting, attribution would treat all activated neurons equally, conflating task-relevant computations with irrelevant background activations.
We evaluate gradient weighting's impact across all 16 configurations in Round-10 (4 model architectures $\times$ 4 domains) from Section~\ref{sec:tool-overall-performance}.
For each configuration, we analyze provenance attribution performance at every individual transformer block layer, computing accuracy under two conditions:
(1) \textit{with gradient weighting}, using the complete \tool method where client contributions are relevance-weighted by token gradients (Equation~\ref{eq:layer-provenance}), and 
(2) \textit{without gradient weighting}, where client contributions are measured using only activation magnitudes, ignoring gradient-based relevance filtering.
For each layer, we use 20 test inputs (after passing trigger test as described earlier) to compute \tool's provenance attribution accuracy. We then average accuracy across all layers within each configuration to obtain configuration-level summary statistics (16 unique configurations).
This per-layer evaluation isolates gradient weighting's effect \emph{by examining attribution at individual layers}, allowing us to precisely measure how gradient weighting enhances layer-wise attribution.

Figure~\ref{fig:gradient-ablation-bar} presents the averaged per-layer attribution accuracy for each configuration under both conditions.
Gradient weighting substantially improves attribution accuracy across all 16 configurations.
With gradient weighting enabled, \tool achieves a mean accuracy of 66.34\% (range: 55.0\%--83.33\%) across all configurations.
When gradient weighting is disabled, mean accuracy drops to 35.71\% (range: 22.94\%--56.20\%), representing an overall 1.86$\times$ improvement from gradient weighting.
Critically, gradient weighting provides consistent improvements across diverse model architectures and domains. These improvements factors across architectures suggest that gradient weighting's effectiveness depends on how models distribute task-relevant information across layers, with some architectures benefiting more from explicit relevance filtering than others.
Notably, even without gradient weighting, \tool maintains non-trivial attribution accuracy in most configurations (22.94\%--56.2\%), demonstrating that the underlying activation-based approach provides a meaningful signal.
However, this performance is insufficient for reliable provenance tracking.
These results validate \tool's core design principle of gradient-based relevance weighting.
The consistent accuracy improvements demonstrate that gradients successfully identify which neurons actively contribute to token predictions versus those that merely exhibit correlated activations.
Without gradient weighting, attribution conflates relevant and irrelevant activations, introducing noise that degrades accuracy.
By weighting each neuron's contribution by its gradient magnitude, \tool automatically filters this noise, focusing attribution on neurons that causally influence the generated token.

  \textbf{Takeaway.} Gradient weighting enhances \tool's attribution accuracy, providing an average 1.86$\times$ improvement (66.34\% vs. 35.71\%) across 16 configurations. 
  While \tool remains functional without gradients, gradient weighting is essential for reliable client(s) attributions. 

\subsection{RQ3: Computational Tractability}
 Naively attributing contributions across all neurons and layers is infeasible (e.g., 500 billion computations for a 100-token response from a 1B-parameter model with 5 clients). \tool addresses this by monitoring only the last $N$ transformer blocks, where task-specific knowledge concentrates~\cite{olah2018building, yu2022spatl}. We measure \tool's overhead and attribution accuracy as we vary the number of monitored layers, from the last 3 layers up to nearly all layers, across four model architectures. Experiments use 5 test samples per global LLM at round 10.
\begin{figure}[t]
  \centering
  \includegraphics[width=0.8\linewidth]{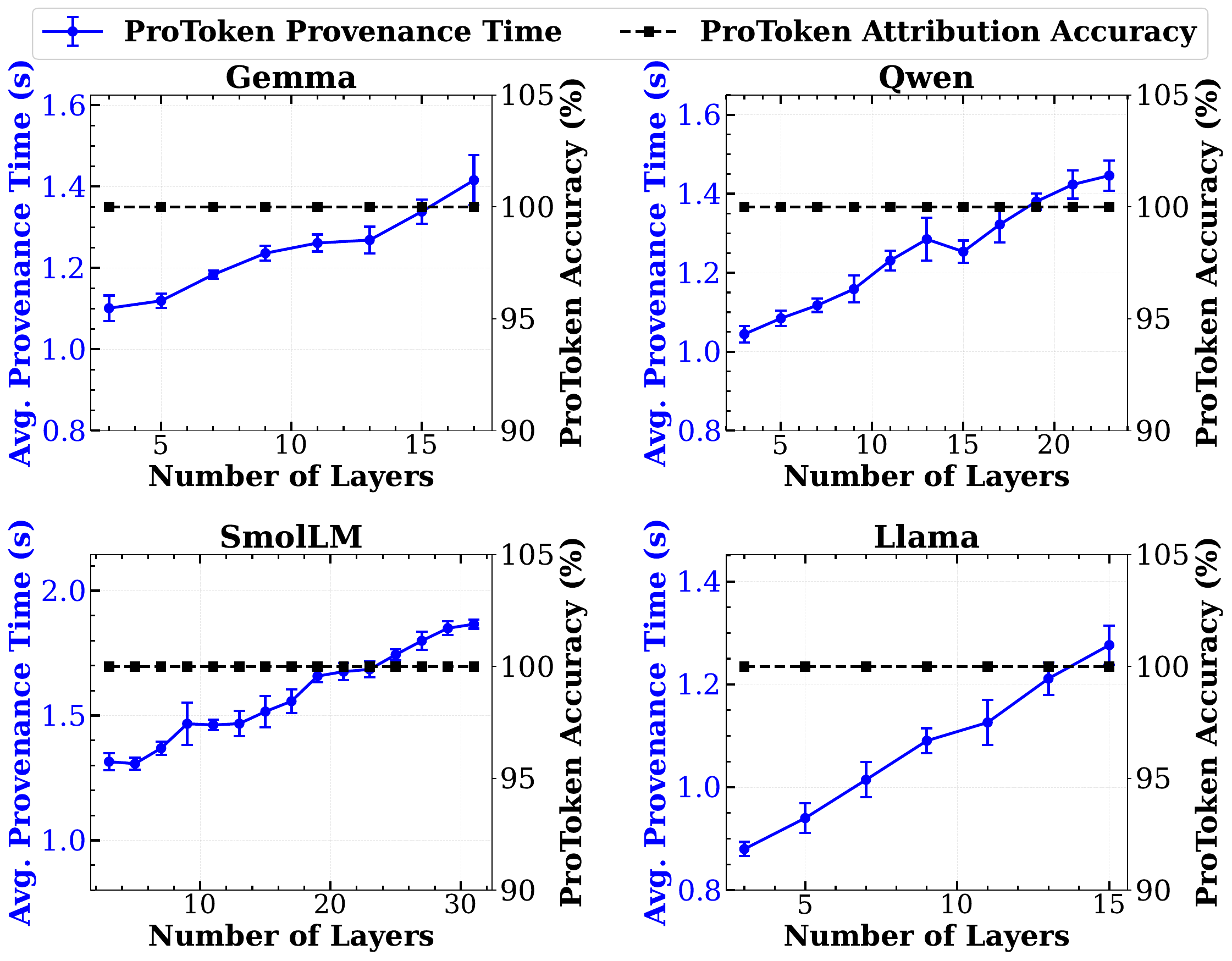}
  \caption{
    For each model, we vary the number of monitored layers (x-axis) and measure \tool's average provenance computation time (left y-axis, blue) and attribution accuracy (right y-axis).
  }
  \label{fig:overhead}
  \vspace{-2em}
\end{figure}

Figure~\ref{fig:overhead} shows that \tool achieves 100\% attribution accuracy for all models and layer counts, confirming that provenance signals are concentrated in later transformer blocks.  For Gemma-3-270M-it (18 total layers), \tool's overhead ranges from 1.10s with 3 layers to 1.42s with last layers, representing a 29\% increase.
Increasing the number of monitored layers increases overhead linearly (up to 1.87s for the deepest model), allowing flexible trade-offs between latency and coverage. 
This validates \tool's efficient design: focusing on key layers enables accurate, scalable provenance tracking without redundant computation. 
Compared to tracking all parameters~\cite{tracefl}, \tool's approach reduces computation by orders of magnitude, making federated LLM provenance tracking viable in practice.

  \textbf{Takeaway.} \tool provides accurate and efficient attribution for federated LLMs by enabling provenance tracking on only the last subset of model layers.

\subsection{RQ4: Scalability Analysis}
While Section~\ref{sec:tool-overall-performance} demonstrated \tool's effectiveness with 6 clients, real-world federated deployments in healthcare, financial networks, and collaborative research often involve dozens of participating organizations.
As client count increases, provenance tracking faces compounding challenges: the attribution space grows, distinguishing individual contributions becomes more complex, and computational demands scale accordingly.

We evaluate \tool with 55 total clients (9.2$\times$ increase from the baseline), injecting backdoor triggers into 25 malicious clients (clients 0-24) while 30 clients (25-54) remain benign.
Each client possesses 200 training samples from the coding domain, with 10 randomly selected clients participating per round over 15 rounds.
We evaluate Gemma and Qwen architectures using the same backdoor-based methodology as Section~\ref{sec:tool-overall-performance}, where correct attribution requires identifying clients 0-24 as the source.

\begin{figure}[t]
  \centering
  \includegraphics[width=0.98\linewidth]{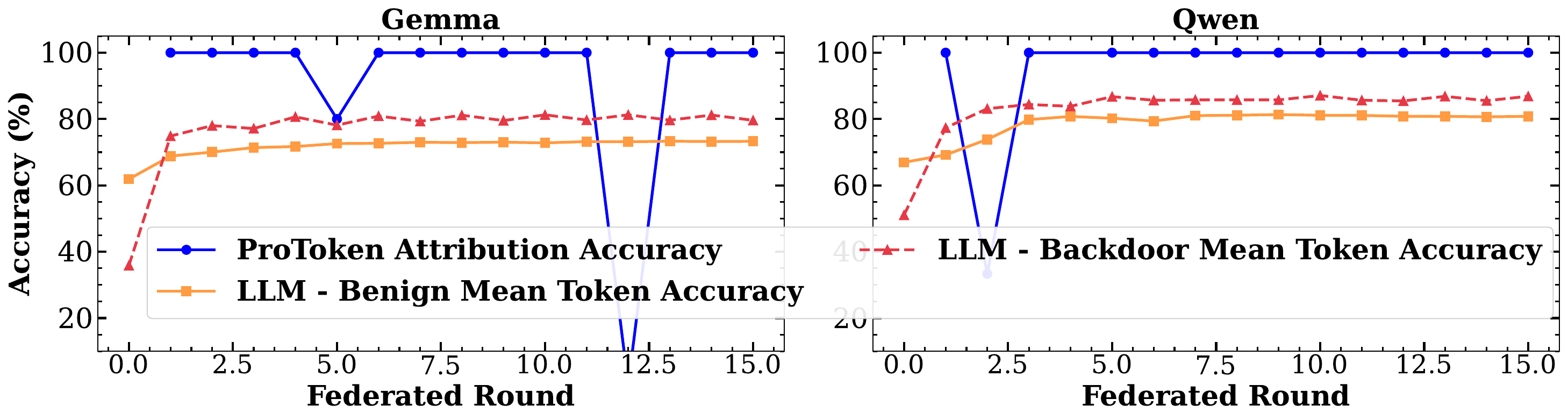}
  \caption{
    \tool maintains high attribution accuracy throughout, demonstrating effective scalability from 6 to 55 clients.
  }
  \label{fig:scalability-results}

\end{figure}

Figure~\ref{fig:scalability-results} shows training dynamics and provenance attribution.
Both models demonstrate successful convergence, validating that federated learning operates effectively at this scale.
For Gemma, benign mean token accuracy improves from 61.89\% at initialization to 73.27\% by round 15 (11.38 percentage point gain), while backdoor accuracy increases from 35.86\% to 79.61\%.
This confirms the model successfully incorporates trigger-response patterns from 25 malicious clients while maintaining benign task performance. \tool maintains high provenance performance despite the 9.2$\times$ increase in client count, achieving 92.00\% average attribution accuracy on Gemma and 95.24\% on Qwen.
Comparing to Section~\ref{sec:tool-overall-performance} (98.62\% with 6 clients, 2 malicious contributors), \tool's performance at 55 clients with 25 malicious contributors represents only modest degradation.
This graceful degradation demonstrates that \tool's core mechanisms scale effectively to larger federated deployments.
\begin{figure}[t]
  \centering
  \includegraphics[width=0.98\linewidth]{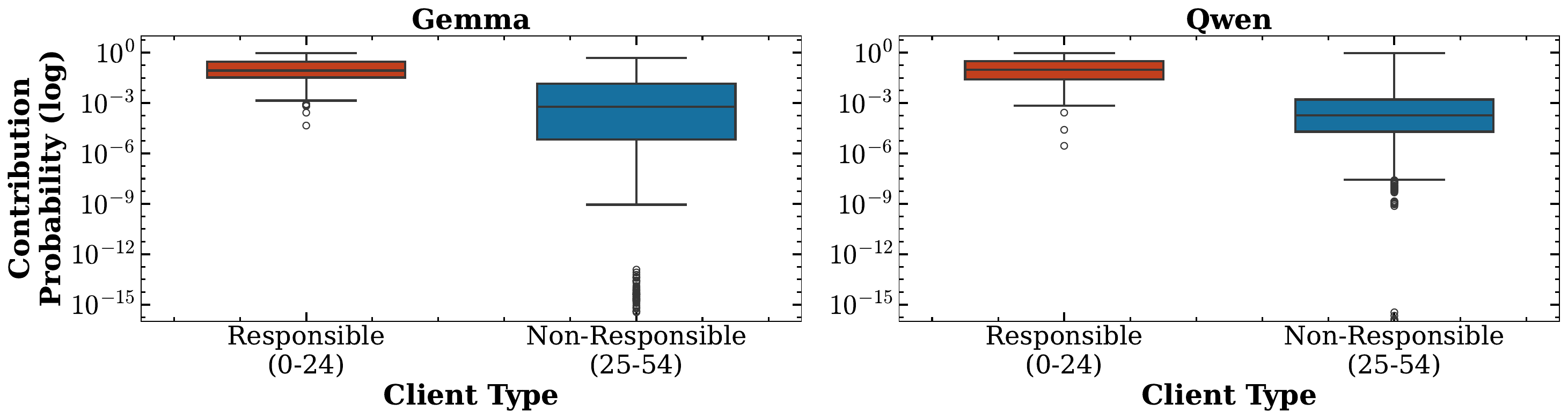}
  \caption{
    \tool maintains clear separation between responsible (0-24) and non-responsible (25-54) clients. 
  }
  \label{fig:scalability-boxplots}
  \vspace{-1em}
\end{figure}
Figure~\ref{fig:scalability-boxplots} presents aggregated client contribution probability distributions.
\tool exhibits clear separation between responsible clients (clients 0-24) and non-responsible clients (clients 25-54), demonstrating that \tool's separation property persists across scales.

\textbf{Takeaway.} \tool successfully scales from 6 clients to 55 clients and maintains high attribution accuracy of more than 92\% with clear probability separation between responsible and non-responsible clients. \tool's core  per token provenance and gradient-based weighting prove robust at scale, validating \tool's practical viability for real-world federated LLM deployments.

%% file: sections/conclusion.tex
\section{Conclusion}

We present \tool, a unique provenance methodology for token-level attribution in federated LLMs that addresses the fundamental challenge of determining which clients contributed to specific generated responses. 
Our comprehensive evaluation across 16 configurations demonstrates that \tool achieves 98.62\% average attribution accuracy and maintains 92-95\% accuracy at scale. These results validate \tool's effectiveness and practical viability for real-world federated LLM deployments, enabling critical applications including debugging, malicious client detection, fair reward allocation, and trust verification in collaborative learning environments.

%% file: main.bib
@inbook{tracefl,
  author    = {Gill, Waris and Anwar, Ali and Gulzar, Muhammad Ali},
  booktitle = {Proceedings of the IEEE/ACM 47th International Conference on Software Engineering},
  isbn      = {9798331505691},
  numpages  = {13},
  pages     = {2264–2276},
  publisher = {IEEE Press},
  title     = {{TraceFL: Interpretability-Driven Debugging in Federated Learning via Neuron Provenance}},
  url       = {https://doi.org/10.1109/ICSE55347.2025.00128},
  year      = {2025}
}

@inproceedings{kuang2024federatedscope,
  title={Federatedscope-llm: A comprehensive package for fine-tuning large language models in federated learning},
  author={Kuang, Weirui and Qian, Bingchen and Li, Zitao and Chen, Daoyuan and Gao, Dawei and Pan, Xuchen and Xie, Yuexiang and Li, Yaliang and Ding, Bolin and Zhou, Jingren},
  booktitle={Proceedings of the 30th ACM SIGKDD Conference on Knowledge Discovery and Data Mining},
  pages={5260--5271},
  year={2024}
}

@article{bender2018data,
  title={Data statements for natural language processing: Toward mitigating system bias and enabling better science},
  author={Bender, Emily M and Friedman, Batya},
  journal={Transactions of the Association for Computational Linguistics},
  volume={6},
  pages={587--604},
  year={2018},
  publisher={MIT Press One Rogers Street, Cambridge, MA 02142-1209, USA journals-info~…}
}

@article{gebru2021datasheets,
  title={Datasheets for datasets},
  author={Gebru, Timnit and Morgenstern, Jamie and Vecchione, Briana and Vaughan, Jennifer Wortman and Wallach, Hanna and Iii, Hal Daum{\'e} and Crawford, Kate},
  journal={Communications of the ACM},
  volume={64},
  number={12},
  pages={86--92},
  year={2021},
  publisher={ACM New York, NY, USA}
}

@ARTICLE{8843893,
  author={Arnold, M. and Bellamy, R. K. E. and Hind, M. and Houde, S. and Mehta, S. and Mojsilović, A. and Nair, R. and Ramamurthy, K. Natesan and Olteanu, A. and Piorkowski, D. and Reimer, D. and Richards, J. and Tsay, J. and Varshney, K. R.},
  journal={IBM Journal of Research and Development}, 
  title={FactSheets: Increasing trust in AI services through supplier's declarations of conformity}, 
  year={2019},
  volume={63},
  number={4/5},
  pages={6:1-6:13},
  doi={10.1147/JRD.2019.2942288}}

@inproceedings{kacianka2021designing,
  title={Designing accountable systems},
  author={Kacianka, Severin and Pretschner, Alexander},
  booktitle={Proceedings of the 2021 ACM conference on fairness, accountability, and transparency},
  pages={424--437},
  year={2021}
}

@inproceedings{sun2023fedbpt,
author = {Sun, Jingwei and Xu, Ziyue and Yin, Hongxu and Yang, Dong and Xu, Daguang and Liu, Yudong and Du, Zhixu and Chen, Yiran and Roth, Holger R.},
title = {FedBPT: efficient federated black-box prompt tuning for large language models},
year = {2024},
publisher = {JMLR.org},
booktitle = {Proceedings of the 41st International Conference on Machine Learning},
articleno = {1919},
numpages = {15},
location = {Vienna, Austria},
series = {ICML'24}
}

@inproceedings{fedbiot,
author = {Wu, Feijie and Li, Zitao and Li, Yaliang and Ding, Bolin and Gao, Jing},
title = {FedBiOT: LLM Local Fine-tuning in Federated Learning without Full Model},
year = {2024},
isbn = {9798400704901},
publisher = {Association for Computing Machinery},
address = {New York, NY, USA},
url = {https://doi.org/10.1145/3637528.3671897},
doi = {10.1145/3637528.3671897},
booktitle = {Proceedings of the 30th ACM SIGKDD Conference on Knowledge Discovery and Data Mining},
pages = {3345–3355},
numpages = {11},
location = {Barcelona, Spain},
series = {KDD '24}
}

@inproceedings{feddebug,
  author    = {Gill, Waris and Anwar, Ali and Gulzar, Muhammad Ali},
  booktitle = {Proceedings of the 45th International Conference on Software Engineering},
  doi       = {10.1109/ICSE48619.2023.00053},
  isbn      = {9781665457019},
  location  = {Melbourne, Victoria, Australia},
  numpages  = {12},
  pages     = {512–523},
  publisher = {IEEE Press},
  series    = {ICSE '23},
  title     = {{FedDebug: Systematic Debugging for Federated Learning Applications}},
  url       = {https://doi.org/10.1109/ICSE48619.2023.00053},
  year      = {2023}
}

@inproceedings{feddefender,
  address   = {New York, NY, USA},
  author    = {Gill, Waris and Anwar, Ali and Gulzar, Muhammad Ali},
  booktitle = {Proceedings of the 1st International Workshop on Dependability and Trustworthiness of Safety-Critical Systems with Machine Learned Components},
  doi       = {10.1145/3617574.3617858},
  isbn      = {9798400703799},
  location  = {San Francisco, CA, USA},
  numpages  = {4},
  pages     = {6–9},
  publisher = {Association for Computing Machinery},
  series    = {SE4SafeML 2023},
  title     = {{FedDefender: Backdoor Attack Defense in Federated Learning}},
  url       = {https://doi.org/10.1145/3617574.3617858},
  year      = {2023}
}

@article{liu2021enabling,
  author     = {Liu, Yejia and Wu, Weiyuan and Flokas, Lampros and Wang, Jiannan and Wu, Eugene},
  doi        = {10.14778/3494124.3494125},
  issn       = {2150-8097},
  issue_date = {November 2021},
  journal    = {Proc. VLDB Endow.},
  month      = nov,
  number     = {3},
  numpages   = {13},
  pages      = {388–400},
  publisher  = {VLDB Endowment},
  title      = {{Enabling SQL-based Training Data Debugging for Federated Learning}},
  url        = {https://doi.org/10.14778/3494124.3494125},
  volume     = {15},
  year       = {2021}
}

@article{kairouz2021advances,
  author    = {Kairouz, Peter and McMahan, H Brendan and Avent, Brendan and Bellet, Aur{\'e}lien and Bennis, Mehdi and Nitin Bhagoji, Arjun and Bonawitz, Kallista and Charles, Zachary and Cormode, Graham and Cummings, Rachel and others},
  journal   = {Foundations and Trends{\textregistered} in Machine Learning},
  number    = {1-2},
  pages     = {1--210},
  publisher = {Now Publishers Boston—Delft},
  title     = {{Advances and Open Problems in Federated Learning}},
  volume    = {14},
  year      = {2021}
}

@inproceedings{yu2022spatl,
  title        = {Spatl: Salient parameter aggregation and transfer learning for heterogeneous federated learning},
  author       = {Yu, Sixing and Nguyen, Phuong and Abebe, Waqwoya and Qian, Wei and Anwar, Ali and Jannesari, Ali},
  booktitle    = {SC22: International Conference for High Performance Computing, Networking, Storage and Analysis},
  pages        = {1--14},
  year         = {2022},
  organization = {IEEE}
}

@inproceedings{mcmahan2017communication,
  title        = {Communication-efficient learning of deep networks from decentralized data},
  author       = {McMahan, Brendan and Moore, Eider and Ramage, Daniel and Hampson, Seth and y Arcas, Blaise Aguera},
  booktitle    = {Artificial intelligence and statistics},
  pages        = {1273--1282},
  year         = {2017},
  organization = {PMLR}
}

@article{olah2018building,
  title   = {The building blocks of interpretability},
  author  = {Olah, Chris and Satyanarayan, Arvind and Johnson, Ian and Carter, Shan and Schubert, Ludwig and Ye, Katherine and Mordvintsev, Alexander},
  journal = {Distill},
  volume  = {3},
  number  = {3},
  pages   = {e10},
  year    = {2018}
}

@article{jiang2020federated,
  title     = {Federated learning in smart city sensing: Challenges and opportunities},
  author    = {Jiang, Ji Chu and Kantarci, Burak and Oktug, Sema and Soyata, Tolga},
  journal   = {Sensors},
  volume    = {20},
  number    = {21},
  pages     = {6230},
  year      = {2020},
  publisher = {Multidisciplinary Digital Publishing Institute}
}

@article{rieke2020future,
  title     = {The future of digital health with federated learning},
  author    = {Rieke, Nicola and Hancox, Jonny and Li, Wenqi and Milletari, Fausto and Roth, Holger R and Albarqouni, Shadi and Bakas, Spyridon and Galtier, Mathieu N and Landman, Bennett A and Maier-Hein, Klaus and others},
  journal   = {NPJ digital medicine},
  volume    = {3},
  number    = {1},
  pages     = {1--7},
  year      = {2020},
  publisher = {Nature Publishing Group}
}

@incollection{long2020federated,
  title     = {Federated learning for open banking},
  author    = {Long, Guodong and Tan, Yue and Jiang, Jing and Zhang, Chengqi},
  booktitle = {Federated learning},
  pages     = {240--254},
  year      = {2020},
  publisher = {Springer}
}

@article{zheng2021applications,
  title     = {Applications of federated learning in smart cities: recent advances, taxonomy, and open challenges},
  author    = {Zheng, Zhaohua and Zhou, Yize and Sun, Yilong and Wang, Zhang and Liu, Boyi and Li, Keqiu},
  journal   = {Connection Science},
  pages     = {1--28},
  year      = {2021},
  publisher = {Taylor \& Francis}
}

@article{lundberg2017unified,
  title   = {A unified approach to interpreting model predictions},
  author  = {Lundberg, Scott M and Lee, Su-In},
  journal = {Advances in neural information processing systems},
  volume  = {30},
  year    = {2017}
}

@inproceedings{shrikumar2017learning,
  title        = {Learning important features through propagating activation differences},
  author       = {Shrikumar, Avanti and Greenside, Peyton and Kundaje, Anshul},
  booktitle    = {International conference on machine learning},
  pages        = {3145--3153},
  year         = {2017},
  organization = {PMLR}
}

@inproceedings{DBLP:journals/corr/SimonyanVZ13,
  author    = {Karen Simonyan and
               Andrea Vedaldi and
               Andrew Zisserman},
  editor    = {Yoshua Bengio and
               Yann LeCun},
  title     = {{Deep Inside Convolutional Networks: Visualising Image Classification
               Models and Saliency Maps}},
  booktitle = {2nd International Conference on Learning Representations, {ICLR} 2014,
               Banff, AB, Canada, April 14-16, 2014, Workshop Track Proceedings},
  year      = {2014},
  bibsource = {dblp computer science bibliography, https://dblp.org}
}

@inproceedings{selvaraju2017grad,
  title     = {Grad-cam: Visual explanations from deep networks via gradient-based localization},
  author    = {Selvaraju, Ramprasaath R and Cogswell, Michael and Das, Abhishek and Vedantam, Ramakrishna and Parikh, Devi and Batra, Dhruv},
  booktitle = {Proceedings of the IEEE international conference on computer vision},
  pages     = {618--626},
  year      = {2017}
}

@inproceedings{zeiler2014visualizing,
  title        = {Visualizing and understanding convolutional networks},
  author       = {Zeiler, Matthew D and Fergus, Rob},
  booktitle    = {Computer Vision--ECCV 2014: 13th European Conference, Zurich, Switzerland, September 6-12, 2014, Proceedings, Part I 13},
  pages        = {818--833},
  year         = {2014},
  organization = {Springer}
}

@inproceedings{ribeiro2016should,
  title     = {" Why should i trust you?" Explaining the predictions of any classifier},
  author    = {Ribeiro, Marco Tulio and Singh, Sameer and Guestrin, Carlos},
  booktitle = {Proceedings of the 22nd ACM SIGKDD international conference on knowledge discovery and data mining},
  pages     = {1135--1144},
  year      = {2016}
}

@inproceedings{sun2022causality,
  title     = {Causality-based neural network repair},
  author    = {Sun, Bing and Sun, Jun and Pham, Long H and Shi, Jie},
  booktitle = {Proceedings of the 44th International Conference on Software Engineering},
  pages     = {338--349},
  year      = {2022}
}

@inproceedings{usman2021nn,
  title        = {NNrepair: Constraint-based repair of neural network classifiers},
  author       = {Usman, Muhammad and Gopinath, Divya and Sun, Youcheng and Noller, Yannic and P{\u{a}}s{\u{a}}reanu, Corina S},
  booktitle    = {Computer Aided Verification: 33rd International Conference, CAV 2021, Virtual Event, July 20--23, 2021, Proceedings, Part I 33},
  pages        = {3--25},
  year         = {2021},
  organization = {Springer}
}

@inproceedings{gerasimou2020importance,
  title     = {Importance-driven deep learning system testing},
  author    = {Gerasimou, Simos and Eniser, Hasan Ferit and Sen, Alper and Cakan, Alper},
  booktitle = {Proceedings of the ACM/IEEE 42nd International Conference on Software Engineering},
  pages     = {702--713},
  year      = {2020}
}

@article{10.1145/3490489,
  author     = {Xie, Xiaofei and Li, Tianlin and Wang, Jian and Ma, Lei and Guo, Qing and Juefei-Xu, Felix and Liu, Yang},
  title      = {{NPC: Neuron Path Coverage via Characterizing Decision Logic of Deep Neural Networks}},
  year       = {2022},
  issue_date = {July 2022},
  publisher  = {Association for Computing Machinery},
  address    = {New York, NY, USA},
  volume     = {31},
  number     = {3},
  issn       = {1049-331X},
  doi        = {10.1145/3490489},
  journal    = {ACM Trans. Softw. Eng. Methodol.},
  month      = apr,
  articleno  = {47},
  numpages   = {27}
}

@article{tao2023dlregion,
  title     = {DLRegion: coverage-guided fuzz testing of deep neural networks with region-based neuron selection strategies},
  author    = {Tao, Chuanqi and Tao, Yali and Guo, Hongjing and Huang, Zhiqiu and Sun, Xiaobing},
  journal   = {Information and Software Technology},
  volume    = {162},
  pages     = {107266},
  year      = {2023},
  publisher = {Elsevier}
}

@INPROCEEDINGS{chefer2021transformer,
  author={Chefer, Hila and Gur, Shir and Wolf, Lior},
  booktitle={2021 IEEE/CVF Conference on Computer Vision and Pattern Recognition (CVPR)}, 
  title={Transformer Interpretability Beyond Attention Visualization}, 
  year={2021},
  volume={},
  number={},
  pages={782-791},
  doi={10.1109/CVPR46437.2021.00084}}

@inproceedings{sundararajan2017integrated,
author = {Sundararajan, Mukund and Taly, Ankur and Yan, Qiqi},
title = {Axiomatic attribution for deep networks},
year = {2017},
publisher = {JMLR.org},
booktitle = {Proceedings of the 34th International Conference on Machine Learning - Volume 70},
pages = {3319–3328},
numpages = {10},
location = {Sydney, NSW, Australia},
series = {ICML'17}
}

@inproceedings{achtibat2024attention,
author = {Achtibat, Reduan and Hatefi, Sayed Mohammad Vakilzadeh and Dreyer, Maximilian and Jain, Aakriti and Wiegand, Thomas and Lapuschkin, Sebastian and Samek, Wojciech},
title = {AttnLRP: attention-aware layer-wise relevance propagation for transformers},
year = {2024},
publisher = {JMLR.org},
booktitle = {Proceedings of the 41st International Conference on Machine Learning},
articleno = {6},
numpages = {34},
location = {Vienna, Austria},
series = {ICML'24}
}

@article{rashkin2023measuring,
    title = "Measuring Attribution in Natural Language Generation Models",
    author = "Rashkin, Hannah  and
      Nikolaev, Vitaly  and
      Lamm, Matthew  and
      Aroyo, Lora  and
      Collins, Michael  and
      Das, Dipanjan  and
      Petrov, Slav  and
      Tomar, Gaurav Singh  and
      Turc, Iulia  and
      Reitter, David",
    journal = "Computational Linguistics",
    volume = "49",
    number = "4",
    month = dec,
    year = "2023",
    address = "Cambridge, MA",
    publisher = "MIT Press",
    url = "https://aclanthology.org/2023.cl-4.2/",
    doi = "10.1162/coli_a_00486",
    pages = "777--840",
}

@inproceedings{gao2023enabling,
    title = "Enabling Large Language Models to Generate Text with Citations",
    author = "Gao, Tianyu  and
      Yen, Howard  and
      Yu, Jiatong  and
      Chen, Danqi",
    editor = "Bouamor, Houda  and
      Pino, Juan  and
      Bali, Kalika",
    booktitle = "Proceedings of the 2023 Conference on Empirical Methods in Natural Language Processing",
    month = dec,
    year = "2023",
    address = "Singapore",
    publisher = "Association for Computational Linguistics",
    url = "https://aclanthology.org/2023.emnlp-main.398/",
    doi = "10.18653/v1/2023.emnlp-main.398",
    pages = "6465--6488",
}

@article{gemma,
  publtype={informal},
  author={Aishwarya Kamath and Johan Ferret and Shreya Pathak and Nino Vieillard and Ramona Merhej and Sarah Perrin and Tatiana Matejovicova and Alexandre Ramé and Morgane Rivière and Louis Rouillard and Thomas Mesnard and Geoffrey Cideron and Jean-Bastien Grill and Sabela Ramos and Edouard Yvinec and Michelle Casbon and Etienne Pot and Ivo Penchev and Gaël Liu and Francesco Visin and Kathleen Kenealy and Lucas Beyer and Xiaohai Zhai and Anton Tsitsulin and Róbert Busa-Fekete and Alex Feng and Noveen Sachdeva and Benjamin Coleman and Yi Gao and Basil Mustafa and Iain Barr and Emilio Parisotto and David Tian and Matan Eyal and Colin Cherry and Jan-Thorsten Peter and Danila Sinopalnikov and Surya Bhupatiraju and Rishabh Agarwal and Mehran Kazemi and Dan Malkin and Ravin Kumar and David Vilar and Idan Brusilovsky and Jiaming Luo and Andreas Steiner and Abe Friesen and Abhanshu Sharma and Abheesht Sharma and Adi Mayrav Gilady and Adrian Goedeckemeyer and Alaa Saade and Alexander Kolesnikov and Alexei Bendebury and Alvin Abdagic and Amit Vadi and András György and André Susano Pinto and Anil Das and Ankur Bapna and Antoine Miech and Antoine Yang and Antonia Paterson and Ashish Shenoy and Ayan Chakrabarti and Bilal Piot and Bo Wu and Bobak Shahriari and Bryce Petrini and Charlie Chen and Charline Le Lan and Christopher A. Choquette-Choo and CJ Carey and Cormac Brick and Daniel Deutsch and Danielle Eisenbud and Dee Cattle and Derek Cheng and Dimitris Paparas and Divyashree Shivakumar Sreepathihalli and Doug Reid and Dustin Tran and Dustin Zelle and Eric Noland and Erwin Huizenga and Eugene Kharitonov and Frederick Liu and Gagik Amirkhanyan and Glenn Cameron and Hadi Hashemi and Hanna Klimczak-Plucinska and Harman Singh and Harsh Mehta and Harshal Tushar Lehri and Hussein Hazimeh and Ian Ballantyne and Idan Szpektor and Ivan Nardini},
  title={Gemma 3 Technical Report},
  year={2025},
  month={March},
  cdate={1740787200000},
  journal={CoRR},
  volume={abs/2503.19786},
  url={https://doi.org/10.48550/arXiv.2503.19786}
}

@inproceedings{
smollm,
title={Smol{LM}2: When Smol Goes Big {\textemdash} Data-Centric Training of a Fully Open Small Language Model},
author={Loubna Ben allal and Anton Lozhkov and Elie Bakouch and Gabriel Martin Blazquez and Guilherme Penedo and Lewis Tunstall and Andr{\'e}s Marafioti and Agust{\'\i}n Piqueres Lajar{\'\i}n and Hynek Kydl{\'\i}{\v{c}}ek and Vaibhav Srivastav and Joshua Lochner and Caleb Fahlgren and Xuan Son NGUYEN and Ben Burtenshaw and Cl{\'e}mentine Fourrier and Haojun Zhao and Hugo Larcher and Mathieu Morlon and Cyril Zakka and Colin Raffel and Leandro Von Werra and Thomas Wolf},
booktitle={Second Conference on Language Modeling},
year={2025},
url={https://openreview.net/forum?id=3JiCl2A14H}
}

@misc{llama,
  title={Llama 3.2: Revolutionizing Edge AI and Vision with Open, Customizable Models},
  author={{Meta AI}},
  year={2024},
  howpublished={Meta AI Blog},
  url={https://ai.meta.com/blog/llama-3-2-connect-2024-vision-edge-mobile-devices/}
}

@misc{qwen,
  title={Qwen2.5-LLM: Extending the Boundary of LLMs},
  author={{Qwen Team}},
  year={2024},
  howpublished={Alibaba Cloud Community Blog},
  url={https://www.alibabacloud.com/blog/qwen2-5-llm-extending-the-boundary-of-llms_601786}
}

@inproceedings{flower,
  title={Flower: A Friendly Federated Learning Research Framework},
  author={Beutel, Daniel J. and Topal, Taner and Mathur, Akhil and Qiu, Xinchi and Parcollet, Titouan and Lane, Nicholas D.},
  booktitle={arXiv preprint arXiv:2007.14390},
  year={2020}
}

@inproceedings{huggingface,
  title={Transformers: State-of-the-Art Natural Language Processing},
  author={Wolf, Thomas and Debut, Lysandre and Sanh, Victor and Chaumond, Julien and Delangue, Clement and Moi, Anthony and Cistac, Pierric and Rault, Tim and Louf, R{\'e}mi and Funtowicz, Morgan and Davison, Joe and Shleifer, Sam and von Platen, Patrick and Ma, Clara and Jernite, Yacine and Plu, Julien and Xu, Canwen and Le Scao, Teven and Gugger, Sylvain and Drame, Mariama and Lhoest, Quentin and Rush, Alexander M.},
  booktitle={Proceedings of the 2020 Conference on Empirical Methods in Natural Language Processing: System Demonstrations},
  pages={38--45},
  year={2020},
  publisher={Association for Computational Linguistics},
  url={https://www.aclweb.org/anthology/2020.emnlp-demos.6}
}

@inproceedings{fedss,
title={Fed{SS}: Federated Learning with Smart Selection of Clients},
author={Ammar Tahir and Yongzhou Chen and Prashanti Nilayam},
booktitle={Federated Learning Systems (FLSys) Workshop @ MLSys 2023},
year={2023},
url={https://openreview.net/forum?id=kSIJ3ScQ-e}
}

@inproceedings{oort,
  title={Oort: Efficient federated learning via guided participant selection},
  author={Lai, Fan and Zhu, Xiangfeng and Madhyastha, Harsha V and Chowdhury, Mosharaf},
  booktitle={15th $\{$USENIX$\}$ Symposium on Operating Systems Design and Implementation ($\{$OSDI$\}$ 21)},
  pages={19--35},
  year={2021}
}

@inproceedings{reward,
 author = {Xu, Xinyi and Lyu, Lingjuan and Ma, Xingjun and Miao, Chenglin and Foo, Chuan Sheng and Low, Bryan Kian Hsiang},
 booktitle = {Advances in Neural Information Processing Systems},
 editor = {M. Ranzato and A. Beygelzimer and Y. Dauphin and P.S. Liang and J. Wortman Vaughan},
 pages = {16104--16117},
 publisher = {Curran Associates, Inc.},
 title = {Gradient Driven Rewards to Guarantee Fairness in Collaborative Machine Learning},
 url = {https://proceedings.neurips.cc/paper_files/paper/2021/file/8682cc30db9c025ecd3fee433f8ab54c-Paper.pdf},
 volume = {34},
 year = {2021}
}

@article{yang2023fingpt,
  title={{FinGPT: Open-Source Financial Large Language Models}},
  author={Yang, Hongyang and Liu, Xiao-Yang and Wang, Christina Dan},
  journal={FinLLM at IJCAI},
  year={2023}
}

@article{han2023medalpaca,
  title={{MedAlpaca -- An Open-Source Collection of Medical Conversational AI Models and Training Data}},
  author={Han, Tianyu and Adams, Lisa C and Papaioannou, Jens-Michalis and Grundmann, Paul and Oberhauser, Tom and L{\"o}ser, Alexander and Truhn, Daniel and Bressem, Keno K},
  journal={arXiv preprint arXiv:2304.08247},
  year={2023}
}

@misc{codealpaca,
  author = {Sahil Chaudhary},
  title = {Code Alpaca: An Instruction-following LLaMA model for code generation},
  year = {2023},
  publisher = {GitHub},
  journal = {GitHub repository},
  howpublished = {\url{https://github.com/sahil280114/codealpaca}},
}

@inproceedings{
saxton2018analysing,
title={{Analysing Mathematical Reasoning Abilities of Neural Models}},
author={David Saxton and Edward Grefenstette and Felix Hill and Pushmeet Kohli},
booktitle={International Conference on Learning Representations},
year={2019},
url={https://openreview.net/forum?id=H1gR5iR5FX},
}

@article{gao2025flowertune,
  title={FlowerTune: A Cross-Domain Benchmark for Federated Fine-Tuning of Large Language Models},
  author={Gao, Yan and Scamarcia, Massimo Roberto and Fernandez-Marques, Javier and Naseri, Mohammad and Ng, Chong Shen and Stripelis, Dimitris and Li, Zexi and Shen, Tao and Bai, Jiamu and Chen, Daoyuan and others},
  journal={arXiv preprint arXiv:2506.02961},
  year={2025}
}

@inproceedings{
  li2025backdoorllm,
  title={{Backdoor{LLM}: A Comprehensive Benchmark for Backdoor Attacks and Defenses on Large Language Models}},
  author={Yige Li and Hanxun Huang and Yunhan Zhao and Xingjun Ma and Jun Sun},
  booktitle={The Thirty-ninth Annual Conference on Neural Information Processing Systems Datasets and Benchmarks Track},
  year={2025},
}

@inproceedings{llm-layers1,
    title = "{BERT} Rediscovers the Classical {NLP} Pipeline",
    author = "Tenney, Ian  and
      Das, Dipanjan  and
      Pavlick, Ellie",
    editor = "Korhonen, Anna  and
      Traum, David  and
      M{\`a}rquez, Llu{\'i}s",
    booktitle = "Proceedings of the 57th Annual Meeting of the Association for Computational Linguistics",
    month = jul,
    year = "2019",
    address = "Florence, Italy",
    publisher = "Association for Computational Linguistics",
    url = "https://aclanthology.org/P19-1452/",
    doi = "10.18653/v1/P19-1452",
    pages = "4593--4601",
}

@inproceedings{llm-layers2,
    title = "Dissecting Contextual Word Embeddings: Architecture and Representation",
    author = "Peters, Matthew E.  and
      Neumann, Mark  and
      Zettlemoyer, Luke  and
      Yih, Wen-tau",
    editor = "Riloff, Ellen  and
      Chiang, David  and
      Hockenmaier, Julia  and
      Tsujii, Jun{'}ichi",
    booktitle = "Proceedings of the 2018 Conference on Empirical Methods in Natural Language Processing",
    month = oct # "-" # nov,
    year = "2018",
    address = "Brussels, Belgium",
    publisher = "Association for Computational Linguistics",
    url = "https://aclanthology.org/D18-1179/",
    doi = "10.18653/v1/D18-1179",
    pages = "1499--1509",
}
